\pdfoutput=1
\documentclass[11pt,a4paper]{article}
\usepackage[utf8]{inputenc}

\usepackage{times}
\usepackage{latexsym}
\usepackage{url}

\usepackage{mathtools}
\usepackage{setspace} 
\usepackage{dsfont}
\usepackage{amsfonts}
\usepackage{amssymb}
\usepackage{amsmath}

\usepackage{subcaption}
\usepackage{paralist}
\usepackage{times}
\usepackage{latexsym}
\usepackage{graphicx}
\usepackage{booktabs}
\usepackage{numprint}
\usepackage[T1]{fontenc}
\usepackage{tikz}
\usepackage{url}
\usepackage{pgfplotstable}
\usepackage{titlesec}
\usepackage{color}
\usepackage{lipsum,adjustbox}
\usepackage[font={small}]{caption}
\usetikzlibrary{positioning}
\usepackage{bbm}
\usepackage{multirow}
\usepackage{scrextend}
\usepackage{placeins}
\usepackage{tabstackengine}
\usepackage{arydshln}

\makeatletter
\newcommand{\@BIBLABEL}{\@emptybiblabel}
\newcommand{\@emptybiblabel}[1]{}

\usepackage[final]{hyperref}
\usepackage[hyperref]{conll-2019}
\graphicspath{{./plots/}}
\newcommand{\com}[1]{}

\newenvironment{myequation*}{
	\vspace{-1em}
	\begin{equation*}
}{
	\end{equation*}
	\vspace{-1.2em}
}

\aclfinalcopy
\begin{document}
	\title{Automatically Extracting Challenge Sets for Non-local Phenomena  \\
		in Neural Machine Translation}

	\author{
		Leshem Choshen\textsuperscript{1} and Omri Abend\textsuperscript{1,2} \\
		\textsuperscript{1}School of Computer Science and Engineering,
		\textsuperscript{2} Department of Cognitive Sciences \\
		The Hebrew University of Jerusalem \\
		\texttt{leshem.choshen@mail.huji.ac.il, oabend@cs.huji.ac.il}\\
	}
	\maketitle
	
	\begin{abstract}
		We show that the state-of-the-art Transformer MT model is not biased towards monotonic reordering (unlike previous recurrent neural network models), 
		but that nevertheless, long-distance dependencies remain a challenge for the model. Since most dependencies are short-distance, common evaluation metrics will be little influenced by how well systems perform on them. 
		We therefore propose an automatic approach for extracting challenge sets replete with long-distance dependencies, and argue that evaluation using this methodology provides 
		a complementary perspective on system performance.
		To support our claim, we compile challenge sets for English-German and German-English, which are much larger than any previously released challenge set for MT. 
		The extracted sets are large enough to allow reliable automatic evaluation, which makes the proposed approach a scalable and 
		practical solution for evaluating MT performance on the long-tail of syntactic phenomena.\footnote{Our extracted challenge sets and codebase are found in \url{https://github.com/borgr/auto_challenge_sets}.} 
	\end{abstract}
	
	
	\section{Introduction}
	
	The assumption that proximate source words are more likely to correspond to proximate target words has often 
	been introduced as a bias (henceforth, {\it locality bias}) 
	into statistical MT systems \citep{brown1993mathematics,koehn2003statistical,chiang2005hierarchical}.
        While reordering phenomena, abundant for some language pairs,
        violate this simplifying assumption, it has often proved to
        be a useful inductive bias in practice, especially when complemented with targeted techniques for
        addressing non-monotonic translation \citep[e.g.,][]{och2002statistical, chiang2005hierarchical}.
        For example, if an adjective precedes a noun in one language and modifies it syntactically,
          it is likely that their corresponding words will appear close to each other in the translation ---
          i.e., they may not be immediately adjacent or even in the same order in the translation,
          but it is unlikely that they will be arbitrarily distant from one another.
	
	In the era of Neural Machine Translation (NMT), such biases are implicitly introduced by the sequential nature of the LSTM architecture \cite[see \S\ref{sec:related}]{Bahdanau2015Neural}. 
	The influential Transformer model \cite{vaswani2017attention} replaces the sequential LSTMs with self-attention,
	which does not seem to possess this bias. We show that the default implementation of the Transformer does retain some bias,
	but that it can be relieved by using learned positional embeddings (\S\ref{sec:locality}). 
	
	Long-distance dependencies (LDD) between words and phrases  present a long-standing problem for MT \cite{sennrich2016grammatical},
	as they are generally more difficult to detect (indeed, they pose an ongoing challenge for parsing as well \cite{xu2009using}), and often result in non-monotonic translation if the target
	differs from the source in terms of its word order and lexicalization patterns.
	The Transformer's indifference to the absolute position of the tokens raises the question of whether long-distance
	dependencies are still an open problem.

    We address this question by proposing an automatic method to compile challenge sets for evaluating system performance on LDD (\S\ref{sec:corpora}). We distinguish between two main LDD types:
    (1) reordering LDD, namely cases where source and target words largely correspond to one another but are ordered differently; (2) lexical LDD, where the way a word or a contiguous expression on the target side is translated is dependent on non-adjacent words on the source side.
	
	We define a methodology for extracting both LDD types. For reordering LDD, we build on \citet{birch2011reordering},
	whereas for lexical LDD we compile a list of linguistic phenomena that yield LDD, and use a dependency parser to find instances of these phenomena in the source side of a parallel corpus. 
	As a test case, we apply this method to construct challenge sets (\S\ref{sec:cor_comp}) for German-English and English-German.
	The approach can be easily scaled to other languages for which a good enough parser exists.
	
	Experimenting both with RNN and self-attention NMT architectures, we find that although the 
	latter presents no locality bias, LDD remain challenging.
	Moreover, lexical LDD become increasingly challenging with their distance, suggesting that syntactic distance remains an important determinant of performance in state-of-the-art (SoTA) NMT. 
	
	We conclude that evaluating LDD using targeted challenge sets gives a detailed picture of MT performance,
	and underscores challenges the field has yet to fully address.
	As particular types of LDD are not frequent enough to significantly affect coarse-grained measures, such
	as BLEU \cite{papineni2002bleu} or TER \cite{snover2006study}, our evaluation approach provides a complementary perspective on
	system performance.

	\vspace{-0.1cm}
	\section{Related Work}\label{sec:related}
	\vspace{-0.1cm}
	\subsection{Long-distance Dependencies in MT}
	
	A common architecture for text-to-text generation tasks is the (Bi)LSTM encoder-decoder \citep{Bahdanau2015Neural}. 
	This architecture consists of several LSTM layers for the encoder and the decoder and a thin attention layer connecting them. 
	LSTM is a recurrent network with a state vector it updates. 
	At every step, it discards some of the current and past information and aggregates the rest into the state. Any information about the past comes from this state, which is a learned ``summary'' of the previous states \cite[cf.][]{greff2017lstm}. 
	Hence, for information to reach a certain prediction step, it should be stored and then kept throughout the intermediate steps (tokens). 
	While theoretically information could be kept indefinitely \cite{hochreiter1997long}, practical evidence shows that LSTMs performance decreases 
	with the distance between the trigger and the prediction \cite{linzen2016Assessing,Liu2018LSTMs}, and that they have difficulties generalizing over sequence lengths \cite{suzgun2018evaluating}.
	
	Despite being affected by absolute distances between syntactically dependent tokens \cite{linzen2016Assessing}, LSTMs tend to learn to a certain extent structural information even without being instructed to do so explicitly \cite{Gulordava2018ColorlessGR}. \citet{Futrell2018Do} discuss similar linguistic phenomena to what we discuss 
	in \S\ref{sec:cor_comp}, and show that LSTM encoder-decoder systems handle 
	them better than previous N-gram based systems, despite being profoundly affected by distance. 
	
	Transformer \cite{vaswani2017attention} models are also encoder-decoder, but instead of LSTMs, they use self-attention.
	Self-attention is based on gating all outputs of the previous layer as inputs for the current one; put differently, it aggregates all the input in one step. 
	This approach makes information from all 
	parts of the input sequence equally reachable. While this is not the only architecture with such attributes \cite{Oord2016WaveNetAG}, we focus on it due to its SoTA results for MT \cite{Lakew2018ACO}. The Transformer's use of self-attention inspired other works in related fields \cite{Devlin2018BERTPO}, some of which attributed their performance gains 
	to the model's ability to capture long-range context \cite{muller2018ALT}.
	
	As the Transformer does not aggregate input sequentially, token positions must be represented through other means. 
	For that purpose, the embedding of each input token $W$ is concatenated with an embedding of its position in the source sentence $P$. 
	While positional embeddings can generally be any vectors, two implementations are commonly used \cite{Tebbifakhr2018MultisourceTW, guo2018star}:
	learned positional embeddings (learnedPEs; $P$ is randomly initialized), and sine positional embeddings (SinePEs) defined as:
	
	\vspace{-.4cm}
	\begin{align*}
	P_{(pos,2i)} = sin(pos / 10,000^{2i/dim}) \\
	P_{(pos,2i+1)} = cos(pos / 10,000^{2i/dim})
	\end{align*}
	where $dim$ is the dimension of the embedding.
	\citet{vaswani2017attention}  report that they see no benefit in learnedPEs, and hence use SinePEs, which have much fewer parameters.
	
	Most of the dependencies between words are short. Short-distance linguistic dependencies include some
	of the most common phenomena in language, such as determination, modification by an adjective and compounding.
	For example, 62\% of the dependencies in the standard UD EWT training set \cite{silveira14gold} are between tokens that are up to one word apart.
	It stands to reason that the locality bias is useful in these cases.
	Nevertheless, as system quality improves, rarer, more challenging dependencies become a priority,
	and languages present a countless number of long-distance reordering phenomena \citep{deng2017translation}.
	One example is subject-verb agreement, where a correct translation requires that the verb
	is inflected according to the headword of the subject (e.g., in English ``dogs that ..., bark'', while ``a dog that ..., barks'').
	When translating such cases, a locality bias may impede performance, by biasing the model not  
	to attend to both the subject's head and the main verb (which may be arbitrarily distant), thereby disallowing
	it to correctly inflect the main verb.
	
	Due to the benefits of the locality bias, it featured prominently in statistical MT, including 
	in the IBM models, where alignments are constrained not to cross too much \cite{brown1993mathematics},
	and in predicting probabilities of reorderings \cite{koehn2003statistical,chiang2005hierarchical}.
	Difficulties in handling LDD have motivated the development of syntax-based MT \cite{yamada2001syntax},
	that can effectively represent reordering at the phrase level, such as when translating between VSO and SOV languages.
	However, syntax-based MT models remain limited in their ability to map between arbitrarily different word orders \cite{sun2009non,xiong2012modeling}.
	For example, reorderings that violate the assumption that the trees form contiguous phrases would be difficult for most such models to capture.
	In the next section (\S \ref{sec:locality}) we show that the Transformer, when implemented with learnedPEs, presents no locality bias, and hence can,
	in principle, learn dependencies between any two positions of the source, and use them at any step during decoding. 
	
	\subsection{MT Evaluation}
	
	With major improvements in system performance, crude assessments of performance are becoming less satisfying, 
	i.e., evaluation metrics do not give an indication on the performance of MT systems on important challenges for the field  \cite{isabelle2018challenge}.
	String-similarity metrics against a reference are known to be partial and coarse-grained aspects 
	of the task \cite{CallisonBurch2006ReevaluationTR}, but are still the common practice in various text generation tasks.
	However, their opaqueness and difficulty to interpret have led to efforts to improve evaluation measures so that they will better reflect the requirements of the task
	\cite{anderson2016spice, sulem2018Semantic, choshen2018usim}, and to increased interest in defining more interpretable and telling measures \cite{lo2011meant,hodosh2013framing,birch2016hume, choshen2018maege}.
	
	A promising path forward is complementing string-similarity evaluation with linguistically meaningful challenge sets. 
	Such sets have the advantage of being interpretable: they test for specific phenomena that are important for humans and are crucial for language understanding.
	Interpretability also means that evaluation artefacts are more likely to be detected earlier.
	So far, such challenge sets were constructed for French-English \cite{Isabelle2017ACS,isabelle2018challenge} 
	and English-Swedish \cite{ahrenberg20018ACS} \footnote{In WMT 2019 English-German phenomena were tested
            with a new corpus, using both human and automatic evaluation. It is not possible, however, to use
            this evaluation outside the competition \cite{avramidis2019linguistic}.}.
        Previous challenge sets were compiled
	by manually searching corpora for specific phenomena of interest (e.g., yes-no questions which are formulated differently 
	in English and French). These corpora are carefully made but are small in size (ten examples per phenomenon), which means that evaluation must be done manually as well. 
	
	As our methodology extracts sentences automatically based on parser output, we are able to compile much larger challenge sets, which allows us to apply standard MT measures
	to each sub-corpus corresponding to a specific phenomenon.
	The methodology is, therefore, more flexible, and can be straightforwardly adapted to accommodate future advances in MT evaluation.
	
	\vspace{-0.1cm}
	\section{Locality in SoTA NMT}\label{sec:locality}
	\vspace{-0.1cm}
	
	In this section we show that encoder-decoder models based on BiLSTM with attention (see \S\ref{sec:related}), do exhibit 
	a locality bias, but that the Transformer, whose encoder is based on self-attention, and in which token position is encoded
	only through learnedPEs, does not present any such bias.

	\subsection{Methodology}\label{sec:loc_methodology}
	
	In order to test whether an NMT system presents a locality bias in a controlled environment, we examine a setting of arbitrary
	absolute order of the source-side tokens.
	In this case, systems that are predisposed towards monotonic decoding are likely to present lower performance, 
	while systems that have no predisposition as to the order of the target side tokens relative to the source-side tokens are not expected to show any change in performance.
	In order to create a controlled setting, where source-side token order is arbitrary, we extract fixed length sentences, and apply the same permutation to all of them. We then train systems with the permuted source-side data (and
	the same target-side data), and compare results to a control condition where no permutation is applied.
	
	Concretely, we experiment on a German-English setting, extracting all sentences of the most common length (18)
	from the WMT2015 \cite{bojar2015Findings} training data.
	This results in 130,983 sentences, of which we hold out 1,000 sentences for testing. It is comparable in training set size to a low-resource language setting.  
	
	We set a fixed permutation $\sigma:[18] \rightarrow [18]$ and 
	train systems on three versions of the training data (settings): (1) {\sc Regular}, to be used for control; 
	(2) {\sc Permuted} source-side, 
	in which we apply $\sigma$ over all source-side tokens;
	(3) {\sc PerPosEmb} where the positional embeddings of the source-side tokens are permuted;\footnote{
		Formally, if the source sentence is $(t_1,...,t_{18})$, then the input to the Transformer is $\left(\left[W(t_1);P(t_{\sigma(1)})\right],...,\left[W(t_{18});P(t_{\sigma(18)})\right]\right)$.} and (4) {\sc Reversed}, where tokens are input in a reverse order.
	
	We apply the following permutation, $\sigma$, to the source-side tokens:\begin{equation*}
	\small
	\setstacktabbedgap{3.5pt}
	\parenMatrixstack{
		0 & 1 & 2 & 3 & 4 & 5 & 6 & 7 & 8 & 9 & 10 & 11 & 12 & 13 & 14 & 15 & 16 & 17 \\
		11 & 5 & 9 & 15 & 8 & 14 & 10 & 1 & 3 & 16 & 12 & 2 & 0 & 6 & 17 & 4 & 13 & 7
	}
	\end{equation*}
	We did not find any property that would deem this permutation special (examining, e.g., its decomposition into cycles).
	We therefore assume that similar results will hold for other $\sigma$s as well.
	
	We train a Transformer model, optimizing using Adam \cite{kingma2015Adam}. 
	We set the embedding size to 512, dropout rate of 0.1, 6 stack layers in both the encoder 
	and the decoder and 8 attention heads. 
	We use tokenization, truecasing and BPE \cite{sennrich2016neural} as preprocessing, following the same
	protocol as \cite{yang2018improving}. 
	
	We experiment both with learnedPEs, and with SinePEs. 
	We train the BiLSTM model using the Nematus implementation \cite{sennrich2017nematus},
	and use their supplied scripts for preprocessing, training and testing, changing only the datasets used. 
	For all models, we report the highest BLEU score on the test data for any epoch during training, and perform 
	early stopping after 10 consecutive epochs without improvement.
	
	In the Transformer with learnedPEs, 5 repetitions were done in the {\sc Regular} setting,
	and 5 for the other settings: 5 repetitions for {\sc Permuted}, 1 for {\sc PerPosEmb} and 1 for {\sc Reversed}.
	In addition, we trained the BiLSTM model and the Transformer with SinePEs 
	both in the {\sc Regular} condition and in {\sc Permuted}, each was trained once.

	\subsection{Results}
	
	\begin{table}[]
		\resizebox{\columnwidth}{!}{
			\begin{tabular}{@{}llll@{}}
				\toprule
				Model                        & Positional & Setting              & BLEU  \\ \midrule
				\multirow{4}{*}{Transformer}
				& LearntPE     & {\sc Regular}           & 24.81 \\
				& LearntPE     & {\sc Permuted}            & 24.87 (+0.06) \\
				& LearntPE     & {\sc Reverse}           & 24.84 (+0.03) \\
				& LearntPE     & {\sc PerPosEmb} & 24.82 (+0.01) \\
				\multirow{2}{*}{Transformer} & SinePEs      & {\sc Regular}           & 25.08 \\
				& SinePEs      & {\sc Permuted}            & 23.90 (-1.18) \\
				\multirow{2}{*}{Nematus}     &            & {\sc Regular}           & 22.32 \\
				&            & {\sc Permuted}            & 19.67 (-2.65) \\ \cmidrule(l){2-4} 
		\end{tabular}}
		\caption{BLEU score for various Transformer settings on regular and permuted data. In brackets are the differences from {\sc Regular}.
			Nematus and Transformer using SinePEs show decreased performance when permuting the 
			input. Transformer with learnedPEs does not. 
			Rows correspond to  the different models used (Model), which 
			positional embeddings are fed to the Transformer (Positional), and the order of the input tokens (Setting; see text). 
			\label{ta:no_loc} }
		\vspace{-0.3cm}
	\end{table}
	
	Table \ref{ta:no_loc} presents our results. 
	We find that Nematus BiLSTM suffers substantially from permuting the source-side tokens,
	but that the Transformer does not exhibit a locality bias. Indeed, for learnedPEs in all settings ({\sc Regular}, {\sc Permuted}, {\sc Reversed} 
	and {\sc PerPosEmb}), BLEU scores are essentially the same.
	We also find that the common practice of using
	fixed SinePEs does introduce some bias, as attested by the small performance drop between {\sc Regular} and
	{\sc Permuted}.
	Like \citet{vaswani2017attention}, we find that in the {\sc Regular} settings, learnedPEs are not superior  
	in performance to SinePEs, despite having more expressive power. 
	However, our results suggest that the decision between learnedPEs and SinePEs is not without consequences: learnedPEs are preferable if a locality bias is undesired (this is potentially the case for highly divergent language pairs).

	\subsection{Discussion}\label{sec:loc_disc}
	
	Finding that Transformers do not present a locality bias has implications on how to construct their input in MT settings, as well as in other tasks that use self-attention encoders, such as image captioning \cite{you2016image}. 
	It is common practice to augment the source-side with globally-applicable information, e.g., the target language in multi-lingual MT \cite{Johnson2017GooglesMN}. 
	Having no locality bias implies this additional information can be added at any fixed point in the sequence fed to a Transformer, 
	provided that the positional embeddings do not themselves introduce such a bias.
	This is not the case with BiLSTMs, which often require introducing the same information at each input token 
	to allow them to be effectively used by the system \cite{Yao2017BoostingIC,rennie2017self}.

	\vspace{-0.1cm}
	\section{LDD Challenge Sets}\label{sec:corpora}
	\vspace{-0.1cm}
	
	One of the stated motivations of the Transformer model is to effectively tackle long-distance dependencies, which are 
	``a key challenge in many sequence transduction tasks'' \citep{vaswani2017attention}.
	Our results from the previous section show that indeed fixed reordering patterns are completely transparent for
	Transformers. This, however, still leaves the question of how Transformers handle linguistic reordering patterns,
	which may involve varying distances between dependent tokens.

        \subsection{Methodology}
        
	We propose a method for scalably compiling challenge sets to support fine-grained MT evaluation for different
	types of LDD.
	We address two main types:
	
	\paragraph{Reordering LDD} are cases where the words on the two sides of the parallel corpus largely correspond to one another, but are ordered differently.
	These cases may require attending to source words in a highly non-monotonic order, but 
	the generation of each target word is localized to a specific region in the source sentence.
    For example, in English-German, the verb in a subordinated clause appears in a final position,
	  while the verb in the English source appears right after the subject. Consider ``The man that is sitting on the chair'',
	  and the corresponding German ``Der Mann, der auf dem Stuhl sitzt'' (lit. {\it the man, that on the chair sits}) --- while
	  the verb is placed at different clause positions in the two cases, the words mostly have direct correspondents.
	Our methodology follows \citet{birch2011reordering} in detecting such phenomena based on alignment.
	Concretely, we extract a word alignment between corresponding sentences, and collect all sentences that include a pair of aligned words in the
	source and target sides, whose indices have a difference of at least $d \in \mathbb{N}$.
	
	\paragraph{Lexical LDD} are cases where the translation of a single word or phrase is determined by non-adjacent words on the source side.
	This requires attending to two or more regions that can be arbitrarily distant from one another.
	Several phenomena, such as light verbs \cite{isabelle2018challenge},
	are known from the linguistic and MT literature to yield lexical LDD.
	Our methodology takes a predefined set of such phenomena, and defines rules for detecting
	each of them over dependency parses of the source-side.
	See \S\ref{sec:cor_comp} for the list of phenomena we experiment on in this paper.

	Focusing on LDD, we restrict ourselves to instances where the absolute distance between the word and the dependent is at
	least $d\in \mathbb{N}$. 
	Selecting large enough $d$ entails that the extracted phenomena are unlikely to be memorized as a phrase with a specific meaning
	(e.g., encode ``\textit{make} the whole thing \textit{up}'' [$d=3$] as a phrase, rather than as a discontiguous phrase ``make ... up'' with an argument ``the whole thing'').
	This increases the probability that such cases, if translated correctly, reflect the MT systems' ability to recognize that such discontiguous
	units are likely to be translated as a single piece.
	
	We note, that by extracting the challenge set based on syntactic parses, 
	we by no means assume these representations are internally represented by the MT systems in any way, or assume such a representation is required for succeeding in correctly translating such constructions. 
	The extraction method is merely a way of finding phenomena we have a reason to believe are difficult to translate, and meaningful for language understanding. 
	We use Universal Dependencies \cite[UD;][]{nivre2016universal} as a syntactic representation, due to its cross-lingual consistency
	(about 90 languages are supported so far), which allows research on difficult LDD phenomena
	that recur across languages.

	Our extraction methods resemble previous challenge set approaches \citep{Isabelle2017ACS,isabelle2018challenge,ahrenberg20018ACS}, in using linguistically motivated 
	sets of sentence pairs to assess translation quality.  However, as our extraction method is fully automatic, 
	it allows for the compilation of much larger challenge sets over many language pairs. The challenge sets we extract contain hundreds or thousands of pairs (\S \ref{sec:cor_comp}).
	The size of the sets allows using any MT evaluation measures to measure performance,
	and is thus a much more scalable solution than manual inspection, as is commonly done in challenge set approaches.
	
	On the other hand, an automatic methodology has the side-effect of being noisier, and not necessarily selecting
	the most representative sentences for each phenomenon. For instance \textit{befinden sich} (lit. {\it to determine}) 
	includes a verb and a reflexive pronoun, which do not necessarily appear contiguously in German.
	However, as \textit{befinden} always appears with the reflexive \textit{sich}, it might not pose a challenge to NMT systems, which can essentially
	ignore the reflexive pronoun upon translation.

	\subsection{A Test Case on Extracting Sets}\label{sec:cor_comp}
	
	Next, we discuss the compilation of German-English and English-German corpora. 
	We select these pairs, as they are among the most studied in MT, and comparatively high results are obtained for them \cite{bojar2017findings}. Hence, they are more likely to benefit from a fine-grained analysis. 
	
	For the reordering LDD corpus, we align each source and target sentences using FastAlign \cite{dyer2013simple} and collect all sentences 
	with at least one pair of source-side and target-side tokens, whose indices have a difference of at least $d=5$.
	
	\noindent For example:
	\vspace{.1cm}
	
	\begin{addmargin}[1em]{1em}
		\textbf{Source:} \textit{Wäre es ein großer Misserfolg, nicht den Titel in der Ligue 1 zu gewinnen, wie dies in der letzten Saison der Fall war?}    
		
		\noindent \textbf{Gloss:} \textit{Would-be it a big failure, not the title in the Ligue 1 to win, as this in the last season the case was?}
		
		\noindent\textbf{Target:} \textit{In Ligue 1, would not winning the title, like last season, be a big failure?}
	\end{addmargin}
	\vspace{.1cm}
	
	We extract lexical LDD using simple rules over source-side parse trees, parsed with UDPipe \cite{milan2017udpipe}.
	For a sentence to be selected, at least one word should separate the detected pair of words. 
	We picked several well-known challenging constructions for translation that involve discontiguous phrases: 
	reflexive-verb, verb-particle constructions and preposition stranding.
	We note that while these constructions often yield lexical LDDs, and are thus expected to be challenging on average,
	some of their instances can be translated literally (e.g., {\it amuse oneself} is translated to {\it am\"{u}sieren sich}).
	
	\paragraph{Reflexive Verbs.}
	Prototypically, reflexivity is the case where the subject and object corefer.
	Reflexive pronouns in English end with \textit{self} or \textit{selves} (e.g., yourselves) and in German include \textit{sich}, \textit{dich}, \textit{mich} and \textit{uns} among others. 
	However, reflexive pronouns can often change the meaning of a verb unpredictably, and may thus lead to different translations for
	non-reflexive instances of a verb, compared to reflexive ones.
	For example, \textit{abheben} in German means taking off (as of a plane), 
	but \textit{sich abheben} means standing out. 
	Similarly, in the example below, \textit{drängte sich} translates to {\it intrude}, while \textit{dr\"{a}ngte} normally translates to {\it pushed}.
	
	A source sentence is said to include a reflexive verb if one of its tokens is parsed with a reflexive morphological feature (\texttt{refl=yes}).

	\noindent For example:      
	
	\begin{addmargin}[1em]{1em}
		\textbf{Source:} \textit{[...] es ertragen zu müssen, daß eine unsympathische Fremde \textbf{sich} unaufhörlich in ihren Familienkreis \textbf{drängte}.}    \\
		\noindent\textbf{Target:} \textit{[...] to see an uncongenial alien permanently \textbf{intruded} on her own family group.}
	\end{addmargin}

	\paragraph{Phrasal Verbs} are verbs that are made up of a verb and a particle (or several particles), which may change the meaning of the verb unpredictably. 
	Examples of English phrasal verbs include \textit{run into} (in the sense of \textit{meet}) and \textit{give in}, and in German they 
	include examples such as \textit{einladen} ({\it invite}), consisting morphologically of the particle {\it ein} and the verb {\it laden} ({\it load}).
	
	A source sentence is said to include a phrasal verb if a particle dependent (UD labels of \texttt{compound:prt} or \texttt{prt}) exists in the parse.
	{\it trat} in itself means {\it stepped}, but in the extracted example below, {\it trat\ldots entgegen} translates to {\it received}.
	
	\noindent For example:      
	\begin{addmargin}[1em]{1em}
		\textbf{Source:} \textit{[...] ich {\bf trat} ihm in wahnsinniger Wut {\bf entgegen}.}\\ 
		\noindent\textbf{Target:} \textit{[...] I {\bf received} him in frantic sort.}
	\end{addmargin}
	
	\vspace{-0.1cm}
	\paragraph{Preposition Stranding} is the case where a preposition does not appear adjacent to the object it refers to.
	In English, it will often appear at the end of the sentence or a clause. For example, \textit{The banana she stepped \textbf{on}} or \textit{The boy I read the book \textbf{to}}.
	Preposition stranding is common in English and other languages such as Scandinavian languages or Dutch \cite{hornstein1981case}. However, in German, it is not a part of standard 
	written language \cite{beermann2005preposition}, although it does (rarely) appear \cite{fanselow1983einigen}.  
	We, therefore, extract this challenge set only with English as the source side.
	
	While preposition stranding is often regarded as a syntactic phenomenon, we consider it here a lexical LDD, since the translation of prepositions (and in some cases their 
	accompanying verbs) is dependent on the prepositional object, which in the case of preposition stranding, may be distant from the preposition itself. For example, 
	translating {\it the {\bf car} we looked {\bf for}} into German usually uses the verb {\it suchen} ({\it search}), while translating {\it the {\bf car} we looked {\bf at}} does not.
	Translating prepositions is difficult in general \citep{hashemi2014comparison}, but preposition stranding is especially so, as there is no adjacent object to assist disambiguation.
	
	A source sentence is said to include preposition stranding 
	if it contains two nodes with an edge of the type \texttt{obl} (oblique) or a subcategory thereof between them, 
	and the UD POS tag of the dependent is adposition (ADP).
	
	\noindent For example,
	
	\begin{addmargin}[1em]{1em}
		\textbf{Source:} \textit{[...] wherever she wanted to {\bf send} the hedgehog {\bf to} [...]}    \\
		\noindent \textbf{Gloss:} \textit{[...] where she the hedgehog rolled-towards wanted [...]}\\
		\noindent\textbf{Target:} \textit{[...] wo sie den Igel hinrollen wollte [...]}
	\end{addmargin}
	
	\vspace{-0.1cm}
	\subsection{Experiments}\label{sec:chal_exp}
	\vspace{-0.1cm}
	
		\begin{table}[t]
		\centering
		\resizebox{1.\columnwidth}{!}{%
			\begin{tabular}{@{}llll@{}}
				                 & Phenomena             &Books         & Newstest2013 \\
				\multirow{2}{*}{De$\leftrightarrow$En}   &  Reorder         & 7,457    &    306 \\
				& Baseline (full dataset)                 & 51,467       & 3,000\\ 
			\end{tabular}
		}
		\caption{Sizes of reordering and baseline corpora. \label{ta:reord_sizes}}
	\end{table}
	\begin{table}[t]
	\centering
	\resizebox{1.\columnwidth}{!}{%
		\begin{tabular}{@{}lllllll@{}}
			\toprule
			&                       & \multicolumn{4}{c}{Min Distance} &      \\ \midrule
			& Phenomena             &All      & $\geq$1      & $\geq$2      & $\geq$3     & News \\
			\multirow{2}{*}{De$\rightarrow$En}  & Particle              & 8,361    & 7,584   & 6,261  & 4,780  & 232  \\
			& Reflexive             & 13,207   & 8,122   & 5,598  & 4,226  & 281  \\\cmidrule(l){2-7}
			\multirow{3}{*}{En$\rightarrow$De} & Particle              & 4,636    & 786    & 111   & 36    & 17   \\
			& Reflexive             & 3,225    & 1,188   & 460   & 274   & 11   \\
			& Preposition Stranding & 682     & 191    & 85    & 40    & 8    \\ 
			
		\end{tabular}
	}
	
	\caption{Sizes of Lexical LDD corpora. Challenge sets are partitioned (in order of appearance) by the language pairs, the phenomenon type, and the minimal distance between the head and the dependent. Phenomenon appears in the source. Statistics for the Newstest2013 corpora with miminal distance $\ge1$ are at the rightmost column, the rest are on Books.
		\label{ta:sizes}}
	\vspace{-0.8cm}
\end{table}
	\begin{table}[t]
		\resizebox{\columnwidth}{!}{%
			\begin{tabular}{@{}llcccc@{}}
				\toprule
				&                       & \multicolumn{2}{c}{Transformer} & \multicolumn{2}{c}{Nematus} \\ \midrule
				&                       & Books          & News           & Books        & News         \\
				\multirow{4}{*}{De$\rightarrow$En}  & Baseline              & 9.02           & 28.23          & 16.26        & 26.32        \\
				& Reorder               & 7.16           & 22.68              & 13.88        & 22.73        \\
				& Particle              & 7.52           & 27.46          & 15.41        & 23.98        \\
				& Reflexive             & 8.15           & 27.84          & 14.91        & 27.04        \\ \cmidrule(l){2-6} 
				\multirow{5}{*}{En$\rightarrow$De} & Baseline              & 6.33           & 23.7           & 12.25        & 22.03        \\
				& Reorder               & 4.31           & 19.4           & 9.02         & 20.38        \\
				& Particle              & 5.30            & 17.83          & 9.55         & 16.72        \\
				& Reflexive             & 5.07           & 15.77          & 9.97         & 21.81        \\
				& \begin{tabular}{@{}c@{}}Preposition \\Stranding\end{tabular} & 5.37           & 11.82          & 9.73         & 6.27         \\ \cmidrule(l){2-6} 
			\end{tabular}%
		}
		
		\caption{BLEU scores on the challenge sets. Minimum distance between head and dependent $d\ge1$. 
			A clear, consistent drop from the Baseline (full corpus) score is observed in all cases. 
			The top part of the table corresponds to German-to-English (De$\rightarrow$En) sets, and bottom part to English-to-German (En$\rightarrow$De) sets.
			Within each part, rows correspond to various linguistic phenomena (second column), including reordering LDD (Reorder),
			Verb-Particle Constructions (Particle), Reflexive Verbs (Reflexive) and Preposition Stranding. 
			Columns correspond to the models (Tranformer/Nematus), and the domains (Books/News).\label{ta:perf}}
    \vspace{-.2cm}
	
	\end{table}
	
	\begin{table*}[]
		\centering
		\resizebox{1.5\columnwidth}{!}{%
			\begin{tabular}{@{}llllllll@{}}
				\toprule
				&                          &                       & \multicolumn{4}{c}{BLEU} & \multicolumn{1}{c}{Spearman} \\ 
				& Language                 & Phenomena             & All      & $\geq$1      & $\geq$2      & $\geq$3     &                              \\\midrule
				\multirow{5}{*}{Transformer} & \multirow{2}{*}{German}  & Particle              & 7.56  & 7.52  & 7.5   & 7.49  & -0.96 \\
				&                          & Reflexive             & 8.26  & 8.15  & 8.04  & -     & -1            \\
				& \multirow{3}{*}{English} & Particle              & 4.96  & 5.3   & 4.96  & 6.01  & 0.73  \\
				&                          & Reflexive             & 5.41  & 5.07  & 5.25  & 5.04  & -0.63 \\
				&                          & \begin{tabular}{@{}c@{}}Preposition \\Stranding\end{tabular} & 5     & 5.37  & 4.42  & 4.64  & -0.63 \\
				\multirow{5}{*}{Nematus}     & \multirow{2}{*}{German}  & Particle              & 15.48 & 15.41 & 14.45 & 12.36 & -0.92 \\
				&                          & Reflexive             & 15.27 & 14.91 & 15.13 & 13.14 & -0.80  \\
				& \multirow{3}{*}{English} & Particle              & 10.14 & 9.55  & 9.36  & 8.82  & -0.98 \\
				&                          & Reflexive             & 9.46  & 9.97  & 9.54  & 9.35  & -0.36 \\
				&                          & \begin{tabular}{@{}c@{}}Preposition \\Stranding\end{tabular} & 10.01 & 9.73  & 9.14  & 9.04  & -0.97 \\ \cmidrule(l){3-8} 
			\end{tabular}%
		}
		\caption{The effect of dependency distance for lexical LDDs on SoTA performance . 
			Results are in BLEU over the Books challenge sets. 
			Columns correspond to the minimum distance, where \textit{All} does not restrict distance (control). The rightmost column presents the Spearman correlation of the phenomena's score with the minimum distance used. 
			All correlations but one are highly negative, implying that distance has a negative effect on performance. \label{ta:var_dist}}
		\vspace{-0.3cm}
	\end{table*}
	
	We turn to evaluate SoTA NMT performance on the extracted challenge sets.
	
	\paragraph{Experimental Setup.}
	We trained the Transformer on WMT2015 training data \cite{bojar2015Findings}, for parameters see \S\ref{sec:loc_methodology}. 
	For Nematus we used the non-ensemble pre-trained model from 
	\citep{Sennrich2017TheUO}. 
	Each of the test sets, either a baseline or a challenge sets, for the Transformer and Nematus used a maximum of 10k and 1k sentences per 
	set respectively.\footnote{We subsample a smaller test set for Nematus, since the most competitive model for the language pair requires Theano. As Theano is deprecated for two years now, it cannot run on our GPUs, which entails long inference time.}
	
	Two parallel corpora were used for extracting the challenge sets. One is newstest2013 \cite{bojar2015Findings} from the news domain that is commonly used as a development set
	for English-German. The other is the relatively unused Books corpus \cite{tiedemann2012parallel} from the more challenging domain of literary translation.
	The corpora are of sizes 51K and 3K respectively. For lexical LDD, we took the distance ($d$) between the relevant words to be at least 1,
	meaning there is at least one word separating them. See Tables \ref{ta:reord_sizes}, \ref{ta:sizes} for the sizes of the extracted corpora.
	
	For evaluation, we use the MOSES implementation of BLEU \cite{papineni2002bleu,koehn2007koehn}, and for reordering LDD, also RIBES \citep{isozaki2010automatic}, which focuses on reordering. RIBES measures the correlation of n-gram ranks between the output and the reference, where n-gram appears uniquely and in both. 
		
    \paragraph{Manual Validation.}
	\label{sec:manual_val}
    To assess the ability of our procedure to extract relevant LDDs, we manually analyzed over 180 source German sentences extracted from Books, and 81 English ones including all the instances extracted from News and 45 extracted from Books, where instances are evenly distributed between phenomena and distance of exactly 1,2 or 5. 
    We find that 85\% of German sentences, 87\% of the English News sentences and 86\% of the Books ones indeed contain the target phenomenon. 
    For details of the manual evaluation of the extraction procedure, see Appendix \ref{sec:extraction_procedure}.
    
    	\begin{table}[h]
    	\begin{tabular}{llll}
    		&          & News     & Books    \\
    		\multirow{2}{*}{German}  & Baseline & 0.82 & 0.57 \\
    		& Reorder  & 0.79 & 0.54 \\
    		\multirow{2}{*}{English} & Baseline & 0.79 & 0.56  \\
    		& Reorder  & 0.77 & 0.53
    	\end{tabular}%
    	
    	\caption{RIBES scores on the reordering LDD challenge sets. Sentences extracted as being challenging to reorder are harder for the Transformer (lower score). This trend is consistent with our experiments with BLEU. First column indicates the source language.}
    	\label{tab:ribes}
    \end{table}
    
	\paragraph{Results.}
	Comparison of the overall BLEU scores of the NMT models (Table \ref{ta:perf}) against their performance on the challenge sets, shows that the phenomena are challenging for both models. 
	Both in the small development set of newstest2013 and the large set of Books, the challenge subparts are more challenging across the board. For reordering LDD, we further apply RIBES and find a similar trend: RIBES score is lower for the reorder challenge set than the  baseline (see Table \ref{tab:ribes}).

	In order to confirm that the distance between the head and dependent (the ``length'' of the dependency) is related to the 
	observed performance drop in the case of lexical LDD, we partition each of the challenge sets according to their length ($d$),
	and compare the results to a control condition, where all instances of the phenomena listed in \S\ref{sec:cor_comp} are extracted, including non-LDD instances, i.e.,
	sentences where the head and the dependent are adjacent.
	System performance on the sliced challenge sets (Table \ref{ta:var_dist}) shows that performance indeed decreases with $d$. 
	Results thus indicate that it is not only the presence of the phenomena that make these sets challenging, but that the challenge increases with the distance.
	
	We validate this main finding using manual annotation of German to English cases. Using two annotators (with high agreement between them;  $\kappa$=0.79), we find that the decrease in performance with $d$ is replicated. We measure how many of the detected lexical LDD are correctly translated, ignoring the rest of the source and output, as done in manual challenge set approaches. We find that 60\%, 54\% and 38\% of the cases are translated correctly for $d\in \left(1,2,5\right)$, respectively. This suggests that the extracted phenomena and the distance indeed pose a challenge, and that the automatic metric we use shows the correct trend in these cases. See Appendix \ref{sec:inhouse} for details.

        \paragraph{Discussion.}
	Interestingly, these results hold true for the Transformer despite its indifference to the absolute word order. 
	Therefore, word distance in itself is not what makes such phenomena challenging, contrary to what one 
	might expect from the definition of LDD. It seems then that these phenomena are especially challenging due to
        the non-standard linguistic structure (e.g., syntactic and lexical structure),
	and the varying distances in which LDD manifest themselves. 
	The models, therefore, seem to be unable to learn the linguistic structure underlying these phenomena, which may
	motivate more explicit modelling of linguistic biases into NMT models, as proposed by,
        e.g., \citet{Eriguchi2017Learning} and \citet{song2019semantic}.
	
	We note that our experiments were not designed to compare the performance of BiLSTM and self-attention models.
	We, therefore, do not see the Transformer's inferior performance on Books, relative to Nematus as an indication of the
        general ability of this model in out-of-domain settings. 
	What is evident from the results is that translating Books is a challenge in itself, 
	probably due to the register of the language, and the presence of frequent non-literal translations.
	
	A potential confound is that performance might change with the length of the source in BiLSTMs \citep{Carpuat2013ProceedingsOS,Murray2018CorrectingLB}, in Transformers it was reported to increase \cite{Zhang2018AcceleratingNT}. Length is generally greater in the challenge set than in the full test set, and generally increases with $d$, showing if anything a decrease of performance by length. 	
		To assess whether our corpora are challenging due to a length bias, we randomly sample from Books 1,000 corpora with 1,000, 100 and 10 sentences each. The correlation between their corresponding average length and the Transformers' BLEU score on them was 0.06,0.09 and 0.03 respectively. While this suggests length is not a strong predictor of performance, to verify that difficulty is not a result of the distribution of lengths in the challenge sets we conduct another experiment.
	
		For each challenge set and each value of $d$ (0--3), we sample 100 corpora. For each sentence in a given challenge set, we sample a sentence of no more than a difference of 1 in length. This results in a corpus with a similar length distribution, but sampled from the overall population of Books sentences.
	 	Results show that the BLEU score of the challenge sets  in all German to English cases is lower than any randomly sampled corpus.\footnote{Most sampled corpora actually had better scores than the baseline. We believe this is because very short sentences which are mostly noise, are never sampled.} In the English-German cases, trends are similar, albeit less pronounced. This may be due to the low number of long English sentences, which lead to more homogeneous samples. Overall, results suggest that length is extremely unlikely to be the only cause for the observed trends.
	
	\vspace{-0.1cm}
	\section{Conclusion}
	
	As NMT system performance is constantly improving, more reliable methods for identifying and classifying their failures are needed. 
	Much research effort is therefore devoted to developing more fine-grained and interpretable evaluation methods, 
	including challenge-set approaches. 
	In this paper, we showed that, using a UD parser,
	it is possible to extract challenge sets that are large enough to allow scalable MT
	evaluation of important and challenging phenomena. 
	
	An accumulating body of research is devoted to the ability of modern neural architectures such as LSTMs \cite{linzen2016Assessing}
        and pretrained embeddings \cite{hewitt-manning-2019-structural,liu-etal-2019-linguistic,jawahar-etal-2019-bert}
        to represent linguistic features.
	This paper makes a contribution to this literature in confirming that the Transformer model can indeed be made indifferent to
        the absolute order of the words,
	but also shows that this does not entail that the model can overcome the difficulties of LDD in
        naturalistic data. We may carefully conclude then that despite the remarkable feats of current NMT models,
        inducing linguistic structure in its more evasive and challenging instances is still beyond the reach of state-of-the-art NMT,
        which motivates exploring more linguistically-informed models.
	
	\section{Acknowledgments}
	This work was supported by the Israel Science
	Foundation (grant no. 929/17)
	
	
	\FloatBarrier
	\bibliography{propose}
	\bibliographystyle{acl_natbib}
	
	\appendix
	\part*{Appendix}
	\section{Manual Evaluation of Extraction Procedure}\label{sec:extraction_procedure}
	
	Manual evaluation of the sentences extracted using our procedure was performed using two proficient annotators (authors of this paper), one for each source language. These include 180 source German sentences extracted from Books, and 81 English sentences including all the instances extracted from News and 45 extracted from Books. Within Books, sentences are distributed uniformly across phenomena and $d$ values $d \in \{1,2,5\}$. 
	
	In German, phrasal verbs are detected with high precision: 96\% of the sentences indeed include a phrasal verb LDD. 
	For reflexive verbs, 63\% include reflexive verbs with a distance of at least 1, and two thirds of the remaining cases (25\%) include verbal non-reflexive pronouns with $d\geq 1$ (in German, some pronouns may be used both as reflexive and non-reflexive). While non-reflexive verbal pronouns are not lexical LDDs (as they can mostly be translated word by word), they do challenge the system to disambiguate them from reflexive verbs. Our analysis in the next section shows that the extracted non-reflexive cases present similar trends as the reflexive cases.
	
	In English (Table \ref{tab:English_det}) detection precision is high for reflexive and phrasal verbs. Preposition stranding detection precision is lower. However, wrongly extracted examples mostly involve prepositional objects that are elided or difficult to detect. We therefore consider the difficulties such cases pose as sufficiently similar to the ones posed by preposition stranding.

	\section{Manual MT Performance evaluation}\label{sec:inhouse}
	
	Using the same sample of 180 sentences used for German detection (See Manual Validation in the paper), we analyze the performance of the Transformer using in-house annotators. One annotator (an author of the paper), proficient in English and German, was presented with the German source in which the relevant tokens were marked. The annotator was asked to locate and mark the corresponding part in the English reference. Places in which the gold translation did not contain a translation of the phenomena, usually due to alignment errors in the corpus or complete omission by the human translator, are removed from the analysis. Then, two annotators (a different author and a non-author), proficient in English, were asked to judge whether the Transformer output conveys the meaning marked in the reference. Inter-annotator agreement was computed to be  $\kappa=0.79$. 
	
	Results (Table \ref{tab:analisys}) show a decrease in performance when increasing the distance $d$. With reflexive verbs, this effect is smaller between $d=1$ and $d=2$. However, looking at each category separately (reflexive or non-reflexive pronouns) shows that performance decreases with $d$ in all cases (Table \ref{tab:reflexives}).
	

	\begin{table}[bht]
		\begin{tabular}{lll}
			& News & Books \\
			Reflexive             & 0.91 & 0.87  \\
			Preposition Stranding & 0.75 & 0.60  \\
			Particle              & 0.94 & 1.00 
		\end{tabular}
		\caption{Ratio of extracted sentences that indeed present the target lexical LDD in English. Rows correspond to various lexical LDD  types, and column correspond to corpora.}
		\label{tab:English_det}
	\end{table}

	\begin{table}[bht]
		\begin{tabular}{llll|lll}
			& \multicolumn{3}{c}{Amount} &  \multicolumn{3}{c}{Accuracy} \\
			$d=$& \multicolumn{1}{c}{1}       & \multicolumn{1}{c}{2}       & \multicolumn{1}{c}{5}              & \multicolumn{1}{c}{1}        & \multicolumn{1}{c}{2}       & \multicolumn{1}{c}{5}       \\
			Particle  & 28      & 26      & 26      & 0.68     & 0.58    & 0.42    \\
			Reflexive & 20      & 24      & 14      & 0.50     & 0.50    & 0.29    \\
			All       & 48      & 50      & 40          & 0.60     & 0.54    & 0.38   
		\end{tabular}%
		\caption{Results of manual annotation of translation quality per lexical LDD phenomena in German to English translation with the Transformer. Left: amount of sentences annotated for each type. Right: accuracy (ratio of cases deemed to be correctly translated). Columns correspond to the distance $d$, as judged by the annotators. Numbers reported are after removing extraction errors and disagreements.}
		\label{tab:analisys}
	\end{table}
	\begin{table}[bht]
		\begin{tabular}{lllllll}
			& \multicolumn{3}{l}{Amount}        & \multicolumn{3}{l}{Accuracy} \\
			$d=$ & 1  & 2  & 5                       & 1        & 2       & 5       \\
			Non-reflexive   & 3  & 15 & \multicolumn{1}{l|}{4}  & 0.67     & 0.53    & 0.25    \\
			Reflexive & 17 & 9  & \multicolumn{1}{l|}{10} & 0.47     & 0.44    & 0.30   
		\end{tabular}
		\caption{
			Results of manual annotation of translation quality per sub-type of reflexive verbs in German to English translation with the Transformer. Left: amount of sentences annotated for each type. Right: accuracy (ratio of cases deemed to be correctly translated). Columns correspond to the distance $d$, as judged by the annotators. Numbers reported are after removing extraction errors and disagreements.}
		\label{tab:reflexives}
	\end{table}

\end{document}